\documentclass[conference]{ieeeconf}
\usepackage{amsmath}
\usepackage{xcolor, soul}
\sethlcolor{yellow}
\usepackage{threeparttable}
\usepackage{multirow}
\usepackage{graphicx}
\usepackage{algorithm}
\usepackage{algpseudocode}
\usepackage{subcaption}
\usepackage{tabularx}
\usepackage{xcolor}
\usepackage{pifont} 
\usepackage{booktabs}
\usepackage{multirow}
\usepackage{makecell}
\usepackage{balance}
\usepackage[figurename=Fig.]{caption}
\usepackage{csvsimple}
\usepackage{siunitx}
\usepackage{layouts}
\usepackage{booktabs}

\makeatletter
\let\NAT@parse\undefined
\makeatother
\usepackage{hyperref}
\hypersetup{
    pdfborder={0 0 0},
    colorlinks=true,
    urlcolor=teal,
    linkcolor=teal,
    citecolor=teal,
    bookmarks=true,
    bookmarksnumbered=true
}

\graphicspath{{../}}

\begin{document}

\vspace{-5cm}
\title{DexGrip: Multi-modal Soft Gripper with Dexterous Grasping and In-hand Manipulation Capacity}

\author{Xing Wang$^{1,*}$, Liam Horrigan$^{1,2}$, Josh Pinskier$^{1}$, Ge Shi$^{1}$,  Vinoth Viswanathan$^{1}$, Lois Liow$^{1}$\\Tirthankar Bandyopadhyay$^{1}$, Jen Jen Chung$^{2}$, David Howard$^{1}$}
\maketitle
\begin{abstract}
The ability of robotic grippers to not only grasp but also re-position and re-orient objects in-hand is crucial for achieving versatile, general-purpose manipulation. While recent advances in soft robotic grasping has greatly improved grasp quality and stability, their manipulation capabilities remain under-explored. 
This paper presents the DexGrip, a multi-modal soft robotic gripper for in-hand grasping, re-orientation and manipulation. DexGrip features a 3 Degrees of Freedom (DoFs) active suction palm and 3 active (rotating) grasping surfaces, enabling soft, stable, and dexterous grasping and manipulation without ever needing to re-grasp an object. Uniquely, these features enable complete \SI{360}{\degree} rotation in all three principal axes. We experimentally demonstrate these capabilities across a diverse set of objects and tasks. DexGrip successfully grasped, re-positioned, and re-oriented objects with widely varying stiffnesses, sizes, weights, and surface textures; and effectively manipulated objects that presented significant challenges for existing robotic grippers.
\end{abstract}
\IEEEpeerreviewmaketitle
\footnotetext[1]{ CSIRO Robotics, Data61, CSIRO, Australia}
\footnotetext[2]{ University of Queensland, Australia}

\section{Introduction}


    Robotic grasping employs advanced end-effectors to perform a series of actions, including object picking, holding, manipulation, and placement. This capability is crucial for executing automated robotic tasks across diverse domains, such as industrial packaging \cite{laschi2016soft}, agricultural harvesting \cite{zhou2022intelligent}, underwater fragile sample grasping \cite{gong2021soft}, etc.
    However, executing more complex tasks in these domains often requires object re-grasping and re-positioning, which substantially increases the complexity of robotic manipulators' control system due to the need for additional motion planning and collision avoidance strategies \cite{sintov2018dynamic}.
    
    Harvesting an apple from an orchard, for instance, requires a pre-programmed path planning and grasping sequence. The robotic arm must first navigate through dense branches and leaves, then delicately grasp and twist the fruit to detach it from the tree, and finally withdraw it with the robotic arm \cite{wang2023development}. In industrial packing, re-grasps are usually required to correctly orient the arbitrarily placed objects for downstream assembly or kitting \cite{saccuti2024contact}. 
    Grasping and manipulating objects \textit{in-hand} can significantly enhance the speed and efficiency of object handling, reducing the need for complex manipulator planning. However, this requires end-effector with capabilities far beyond just grasping. Dexterous, multi-modal grippers are needed for in-hand re-positioning re-orientation. They must maintain a stable grasp while enabling relative motion between the gripper and objects.
    
    Soft robotics has traditionally focused on conformable grasping of unknown objects. It leverages the inherent compliance of soft materials to bend around objects, distributing forces around the object and creating a soft, stable grasp by maximising the contact area from material compliance \cite{rus2015design}. 
    However, this comes at the expense of manipulability, as the grippers cannot actively reconfigure to re-position themselves or the object while maintaining the stable grasping pattern.
    Previous research has investigated potential solutions to address this issue and enhance soft robotics' capabilities of dexterous in-hand manipulation. 
    For example, multi-chamber pneumatic actuators have enabled multi-DoF dexterous bending for in-hand object manipulation, at the expense of a bulky and complex pressure control system \cite{graule2022somogym}. 
    Alternatively, strategies based on active surfaces have been explored by stacking an inextensible belt on top of compliant fingers, allowing for the re-orientation of objects while the belt is driven \cite{BACH}. 
    However, few studies have focused on enhancing the dexterity of soft grasping by simultaneously maximizing gripping patterns (multi-modal grasping) and in-hand manipulation capabilities (multi-axis rotation).
    
    In this work, we propose DexGrip, a multi-functional soft robotic gripper for multi-modal grasping and in-hand manipulation. 
    It achieves (i) compliant and conformable soft grasping using compact Fin Ray robotic fingers; (ii) active adhesion and manipulation through its active suction palm; (iii) omni-directional in-hand object manipulation through synchronized active surface (belt) rotation, finger actuation and active palm engagement. 
    Through comprehensive experimental evaluations, we demonstrate its grasping and manipulation capability across various objects with distinct geometry, materials, and textures.
    
    
\section{Related Work} \label{related}
    \begin{figure*}[htbp]
        \centering
        \includegraphics[width=0.9\linewidth]{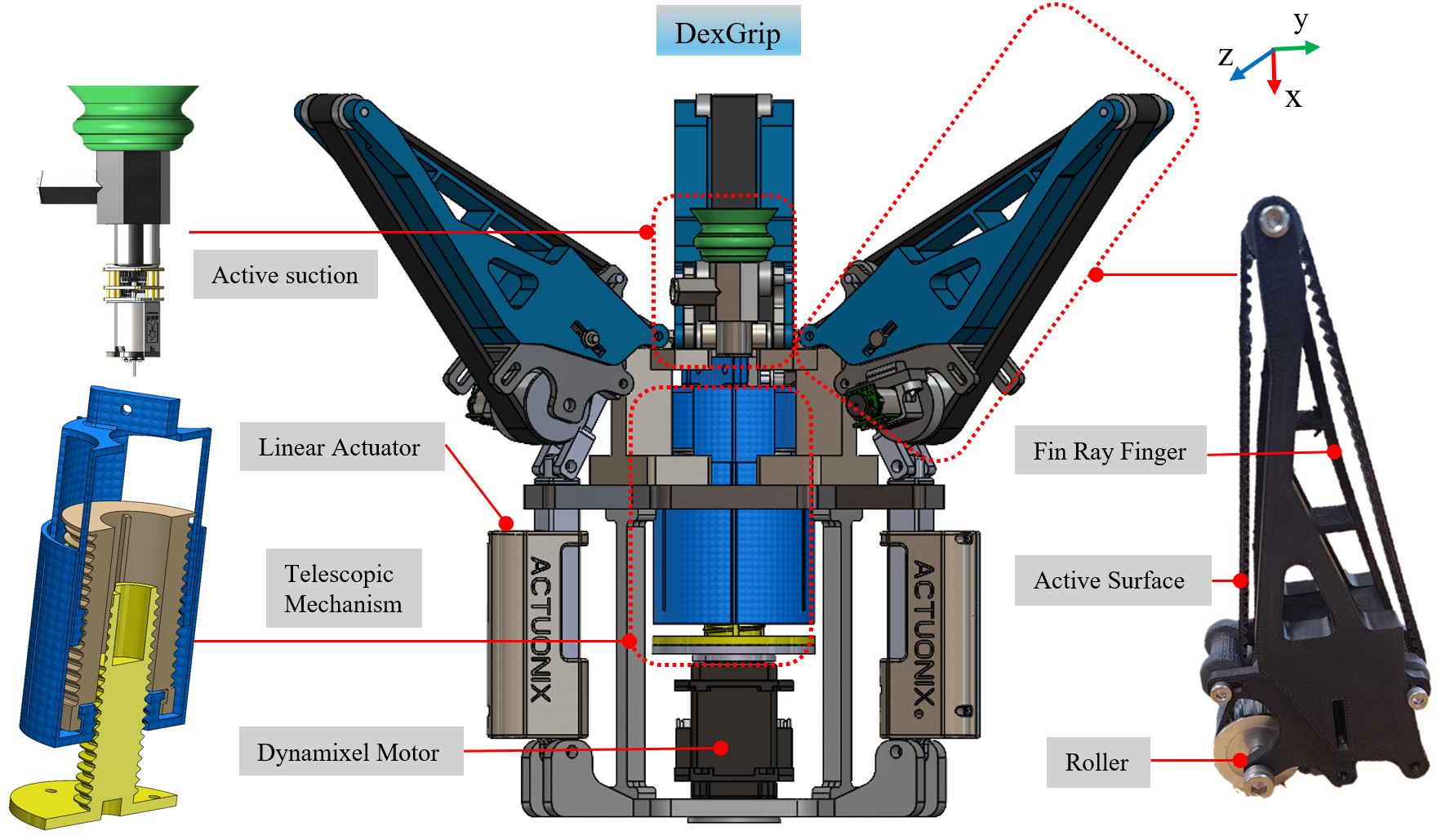}
        \vspace{-0.2cm}        
        \caption{Detailed schematics for the proposed DexGrip design}
        \vspace{-0.4cm}
        \label{fig:design}
    \end{figure*}
  
  Traditional, rigid robotic has made significant progress in developing dexterous grippers for in-hand manipulation. For example, Andrychowicz et al. \cite{andrychowicz2020learning} developed the Shadow Hand, a 24 DOF dexterous hand with a reinforcement learning trained control policy. However, rigid grippers depend heavily on precise and complex planning and control for safe and stable grasping and manipulation. 
  
  Soft grippers can conform to objects without knowing their geometric or material properties\cite{howard2022one,joseph2023jamming, pinskier2024diversity}, making them suitable for manipulating unknown objects. However, it remains a relatively unexplored research area.
  Previous research has explored two approaches for soft in-hand manipulation: one uses multi-chambered pneumatic soft fingers to generate bending in each finger \cite{pagoli2021soft, teeple2021role, zhou2018bcl}; the second incorporates additional active moving components into the hand assembly to create new DOFs independent of the soft finger \cite{BACH, xiang2024adaptive}. 
  The former could facilitate object manipulation through the active control of the tips of pneumatic actuators, while the latter enhances performance by providing supplementary DoFs for under-actuated, soft conforming fingers. These additional DoFs are achieved through mechanisms such as rolling contact points, rotational palm support, and rotating belts.

  As shown in Table \ref{gripper-comparison}, Soft grippers with multi-chamber soft actuators such as balloon actuator \cite{pagoli2021soft, zhou2018bcl}, and PneuNet \cite{teeple2021role}, utilised the soft fingers' multi-directional bending and finger gait to achieve in-hand manipulation capability. 
  However, the indefinite DoFs in these soft materials poses significant challenges for precise and multi-modal control. The complexity of pneumatic control necessitates extensive efforts in physical modelling and parameter tuning to achieve the desired functionality \cite{graule2022somogym}.
  
  Another method employs active surfaces that dynamically adjust the entire contact area to accommodate objects, thereby eliminating the need for finger gaiting to re-position targets. 
  Xiang et al. \cite{xiang2024adaptive} introduced a two-finger active elastic band-based gripper design that enables both translation and rotation of grasped objects through passive adaptive wrapping and active conveyor-driven motion. However, the active surface can only rotate along one principle axis (in a 2D plane), limiting the manipulation functionality.
  Yuan et al. \cite{RollerGrasperV2} developed a robotic gripper featuring pinch grasp and rigid roller mechanisms for object manipulation, which provided enhanced controllability at the expense of stability. 
  In the subsequent work \cite{BACH}, they designed a soft gripper with active surfaces and generated power grasp with limited controllability, which prioritized the grasping stability. However, manipulating the in-hand objects to allow them being secured at different position, or rotate with respect to the central symmetry axis remained to be investigated. 
  Despite the advancements in enabling the manipulation capacity of soft grippers, there is a notable gap in research specifically addressing the integration of active palms and active surfaces to enhance grasping modalities and manipulation dexterity. 
 \begin{table}[htbp]
                \vspace{-0.3cm}
                \caption{Comparison between various grippers }
                \centering
                \resizebox{\columnwidth}{!}{
                    \begin{tabularx}{\columnwidth}
                    {>{\centering\arraybackslash}m{0.5cm} >{\centering\arraybackslash}m{0.5cm}>{\centering\arraybackslash}m{0.7cm}>{\centering\arraybackslash}m{2cm}>{\centering\arraybackslash}m{1.2cm}>{\centering\arraybackslash}m{1.2cm}}
                        \toprule
                        \makecell{Ref} & \makecell{DoFs} & \makecell{Grasping\\Type} & \makecell{Manipulation} & \makecell{Active palm} & \makecell{Collaborative\\Motion} \\
                        \midrule
                        \cite{pagoli2021soft} & 14 & \makecell{Pinch \\ Power} & \makecell{Re-position \\Re-orientation} & \makecell{Suction} &   \ding{51} \\
                        \cite{teeple2021role}  & 5 & \makecell{Pinch } & \makecell{ Re-position\\Re-orientation} & \makecell{-} & \ding{55} \\
                        \cite{zhou2018bcl}  & 13 & \makecell{Pinch\\ } & \makecell{ Re-position\\Re-orientation} & \makecell{-} &  {\centering \ding{55}}\\
                        \cite{xiang2024adaptive}  & 3 & \makecell{Pinch\\Power} & \makecell{Re-position \\Re-orientation} & \makecell{ -} & \ding{55} \\
                        \cite{BACH}  & 4 & \makecell{Pinch\\Power} & \makecell{Re-position \\Re-orientation} & \makecell{-} & \ding{55} \\
                        \makecell{This \\work} & 9 & \makecell{Pinch \\Power} & \makecell{Re-position \\Re-orientation} & \makecell{Suction\\Re-orientation} & \ding{51} \\
                        \bottomrule
                    \end{tabularx}
                }
                \vspace{-0.3cm}
                \label{gripper-comparison}
            \end{table}
  
  To summarise, Table \ref{gripper-comparison} provides a comparison of our work with the state of the art soft grippers that have manipulation capabilities.
    This work surpasses existing designs by offering multiple DoFs, enabling advanced multi-modal grasping (including active palm suction, re-position and re-orientation) and more dexterous (rotation along all 3 principle axes) in-hand manipulation capabilities. To the best of our knowledge, this has not been investigated thoroughly in the existing research. 
    
\section{Methdology} \label{method}
    \begin{figure*}[t]
            \centering
            \includegraphics[width=0.89\linewidth]{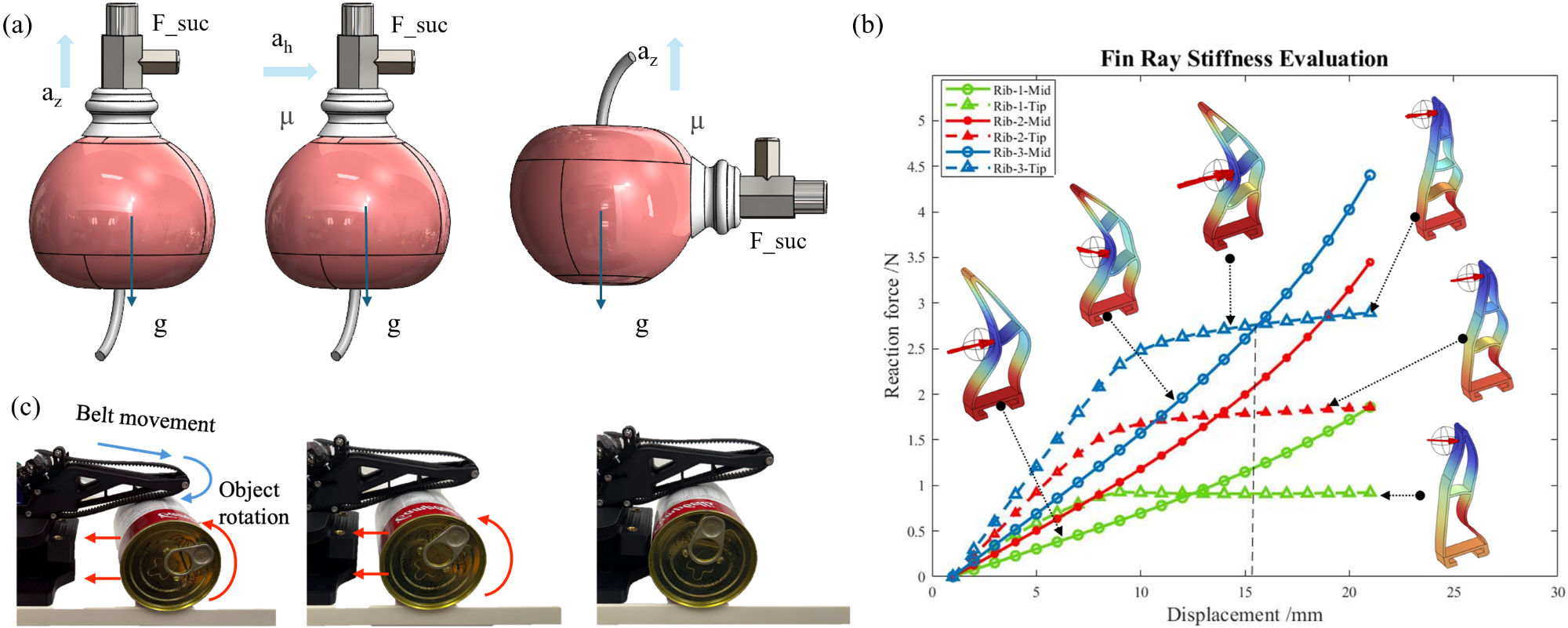}
            \caption{(a) Suction adhesion under three loading cases, (b) Stiffness evaluate for 3 Fin Ray fingers with tip and middle contacts respectively, (c) Single active surface to move/rotate a can against the table. }
            \label{combined-1}
            \vspace{-0.7cm}
        \end{figure*}
    \subsection{Structural Design}

        

        

        

    We developed a prototype of DexGrip with multi-modal grasping and manipulation capabilities, as illustrated in \figurename{} \ref{fig:design}. 
    DexGrip incorporates several critical subsystems. The first component has 3 active surface-augmented Fin Ray fingers with each controlled by an individual linear actuator.
    The Fin Ray fingers would passively conform to targets upon the actuation from linear actuators. The active surface rotates and drives the objects towards the direction of rotation if there is no slippage. 
    The second component is a dedicated active palm that features with an active suction cup for object adhesion and twisting, and a telescopic mechanism driven by a Dynamixel motor for the linear motion.     
    The gripper comprises a total of nine DoFs: three DoFs from the bending of individual fingers, which adjust the grasping range and ensure a secure grasp; three DoFs from the active rotation of the surface on each finger, enabling in-hand manipulation for the soft compliant fingers; and lastly three additional DoFs from the active palm, which enhance grasping ability and facilitate interactions with objects through extension/retraction, suction, and twisting motions. 
    
    \subsection{Subsystem Development}
    
   This section presents the design methodology and analytical process employed for each subsystem.

    \subsubsection{\textbf{Compliant robotic finger}}
    This sub-system requires a complaint finger surface to allow a wide range of objects to be maintained in a grip. We adapt the bio-inspired Fin Ray designs as the baseline owing to their excellent shape adaptation and reasonable force exertion at certain scale \cite{shan2020modeling}.   
    The number of ribs in the baseline design was varied to assess its influence on in-plane bending stiffness, as it is crucial for achieving a secure soft grasping. A high-fidelity finite element method (FEM) solver was integrated within the COMSOL environment to compute the stiffness values. 
    The deformation and reaction force are then recorded to obtain the in-plane bending stiffness. The detailed FEM results regarding the performance characteristics of Fin Ray fingers will be presented in detail in the experimental section. 
    
    Additionally, the finger is equipped with an active in-built belt that facilitates object re-positioning and rotation while ensuring in-plane compliance. The belt allows for the translation and/or rotation of objects, with individual control enabling both clockwise and counterclockwise movement. 
    By applying an appropriate level of tension force, the object is anticipated to move in tandem with the active surfaces, resulting in a similar displacement. This synchronized motion ultimately enables the object's rotation.
     
    \subsubsection{\textbf{Active telescopic palm with adhesion}} 
    An active palm is essential for independently interacting with objects and for collaboratively coordinating with active soft fingers, thereby enhancing the overall manipulation capabilities. It can achieve a sequence of motion such as extension, suction, twisting, and withdraw.
    The design incorporates a suction cup with a single rotational DoF (about x-axis) from the micro DC motor, enabling rotation of the attached object. A telescopic actuator from a Dynamixel servo motor for linear extension and retraction, as shown in \figurename{} \ref{fig:design}. This rotational capability allows the palm alone to manipulate the held object effectively. Consequently, the manipulation can facilitate independent positioning of the object or adjust its orientation to increase contact area with the active surface fingers, resulting in a more stable grasp and expanded DoF for rotation.


    
    The theoretical grasping force of a suction cup can be predicted by taking into account geometric parameters and actuation conditions, providing a framework for estimating the range of objects that can be effectively grasped.
        \begin{align}
            F_{\text{suc}} &= P_{\text{atm}} \cdot A_o - P_{\text{suc}} \cdot A_i \\
            &= P_{\text{atm}} \cdot \frac{\pi (D_o)^2}{4} - P_{\text{suc}} \cdot \frac{\pi (D_i)^2}{4},
            \end{align}

    Where $F_{suc}$ is the suction force, $P_{atm}$ and $P_{suc}$ are the atmosphere, negative vacuum pressure, respectively. $A_o$ and $A_i$ are the outer and inner area of the vacuum cup lip. $D_o$ and $D_i$ are the outer and inner diameter of vacuum cup lip. 

    Considering different loading cases as shown in \figurename{} \ref{combined-1}(a), 
        \begin{equation}
            m_1 =F_{suc} /(g+a_z),
        \end{equation}
        \begin{equation}
            m_2 = F_{suc}/(g+a_h/\mu),
        \end{equation}
        \begin{equation}
            m_3 = \mu\cdot F_{suc}/(g+a_z), 
        \end{equation}
    Where $m_1$, $m_2$, $m_3$ are the feasible object weights for adhesion, $g$ is acceleration of gravity, $\mu$ is the coefficient of friction, and $a_z$ and $a_h$ are the grasp acceleration in the vertical and horizontal directions, respectively.

\section{Experimental Results \& Discussion} \label{experiments}

        

        To validate the performance of the proposed DexGrip, and evaluate its subsystems, we conduct a set of four experiments:
        The first experiment involved testing the stiffness of the Fin Ray structure and its surface manipulation capabilities. This included a FEM-based evaluation of stiffness across three distinct finger designs, and object manipulation enabled by surface rotation.
        The following experiment evaluated the functionality of active palm manipulation, where the active palm was actuated to execute a sequence motion (suction, withdraw, and twisting) on a wide range of objects with distinct features.
        The last experiment evaluated the gripper as a whole assembly. The DexGrip initially was controlled to show its ability to grasp various objects, then to independently perform active surface manipulation. 
        Additionally, the gripper showcased its ability to transition from a pinch grasp to a more stable power grasp. 
        DexGrip also highlighted its unique capability to manipulate objects through the coordinated motion of both the active palm and active surface.

          \subsection{ Fin Ray Stiffness \& Manipulation Testing}
                
            
            Numerical simulations were performed to comprehensively assess the performance of various finger designs. Each finger was subjected to prescribed displacements, making contact with the object at both the fingertip and midsection, representing the contact conditions for pinch and power grasping patterns, as shown in \figurename{} \ref{combined-1} (b).
            It is observed that the contact force increases in both tip and middle contact cases. However, for tip contact, the force shows minimal increase once the input displacement exceeds approximately 10 mm. This occurs because the lateral flat surface of the fingers experiences greater deformation and bulking during tip contact, which compromises the ability to generate sufficient contact force.  
            Furthermore, the middle contact generated a higher force when the input displacement exceeded 10 mm, making it more suitable for stable soft grasping. It is important to note that Fin Ray design optimization was beyond the scope of this work and could be checked in the existing work \cite{wang2024fin, elgeneidy2020structural}.
            The two-ribbed finger configuration was selected for the final DexGrip prototype.
            
            The Fin Ray finger was then integrated with the active surface, and was positioned horizontally and securely fixed at its base, with the target object placed on a table in contact with the active surface, as illustrated in  \figurename{}\ref{combined-1} (c). 
            The active surface was controlled to rotate clockwise through the operation of a Micro DC motor, facilitating both the re-position and rotation of the target.
            
          \subsection{ Active Palm for Re-orientation and Re-position}
                \begin{figure}[b]
                    \centering
                    \vspace{-0.5cm}
                    \includegraphics[width=\linewidth]{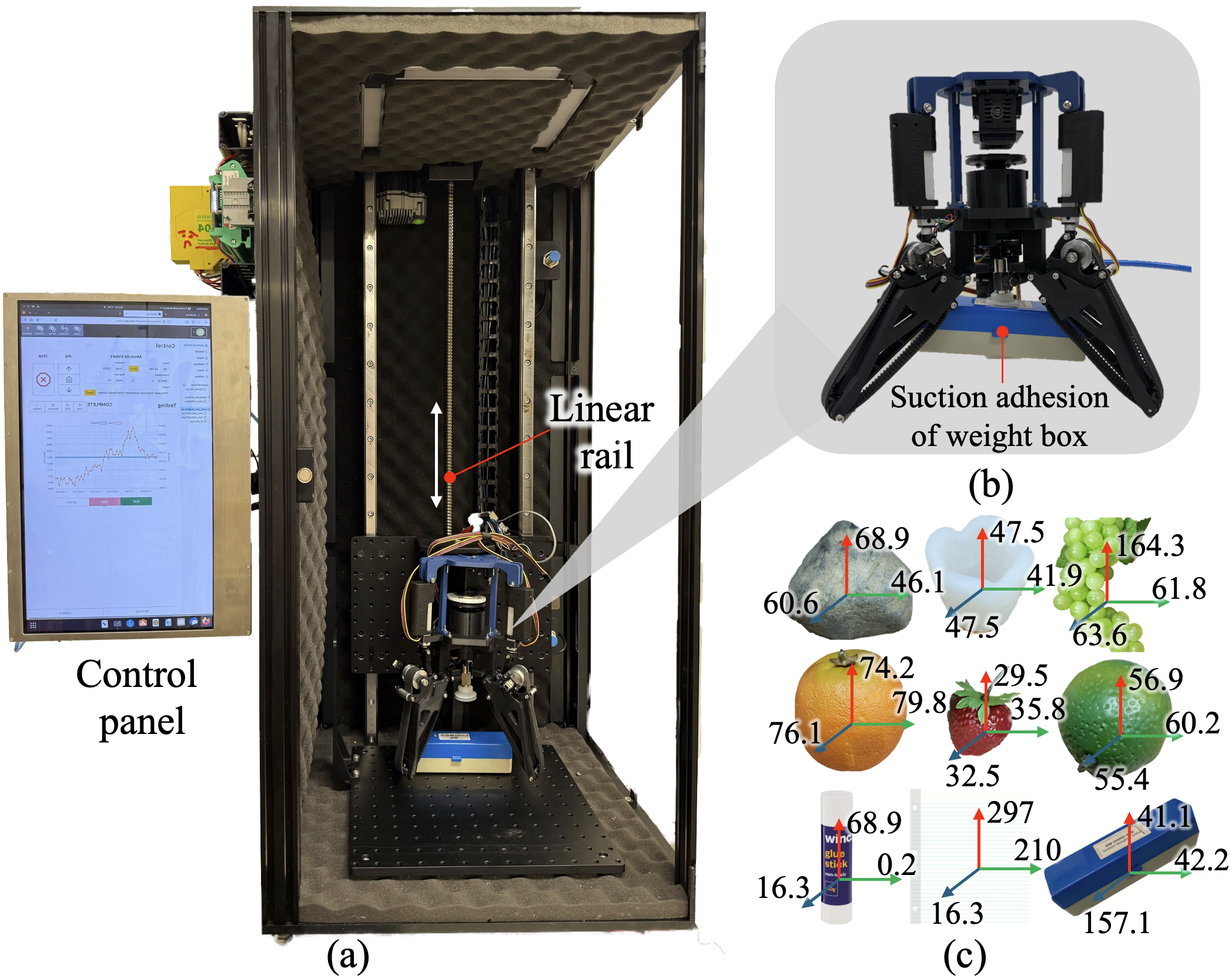}
                    
                    \caption{ (a) Experimental platform (b) zoom in view for DexGrip grasping a weight box, (c) object dataset with detailed dimensions and weights. Note fruits are fake artifacts.}
                    \label{object-image}
                    \vspace{-0.3cm}
                \end{figure}
                In this experiment, the active palm was actuated to execute a sequence of motions. The gripper was placed with the suction cup facing down (\figurename{\ref{object-image}}), it then approached objects from a top-down orientation. The pulling acceleration was minimal and can be considered negligible, as shown in the \textbf{Supplementary Video}.
                Objects with varying geometries, surface textures, stiffnesses, and weights was utilized for this evaluation, as detailed in \figurename{} \ref{object-image} (c). 
                
                \begin{figure}[ht]
                    \centering
                    \includegraphics[width=0.99\linewidth, angle=0]{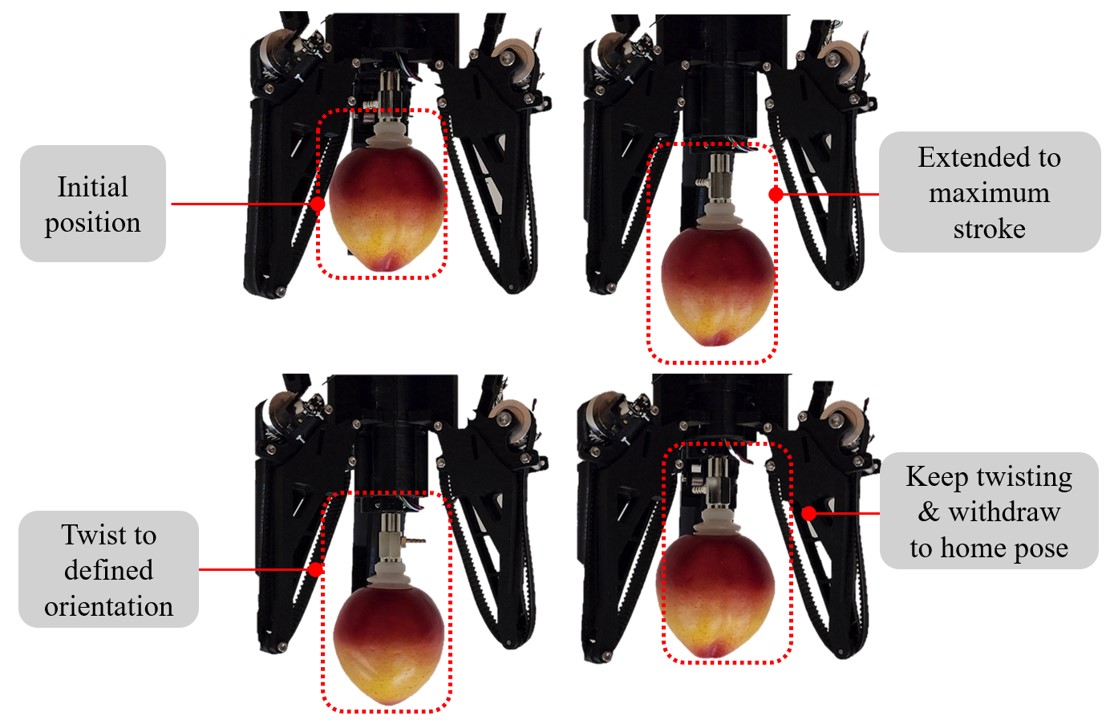}
                    \caption{Active palm executing a sequence of motions on a peach: extend, rotate, and withdraw.}
                    \label{fig: peach}
                    \vspace{-0.9cm}
                \end{figure}

                Aside from the simple grasping, we demonstrated the versatility of our active palm by manipulating a plastic peach, as illustrated in \figurename{} \ref{fig: peach}. The object was initially grasped using the suction force, after which the telescopic actuator positioned it across a continuous range of vertical heights. When re-orientation was necessary, such as when the actuator reached its maximum extension, a twisting motion was employed to align the object at the desired angle, facilitating subsequent manipulation tasks. Ultimately, the object was withdrawn and secured at the center of the palm for further manipulation. 
                
                \begin{table}[t]
                    \vspace{0.1cm}
                    \caption{Experimental results from the active palm}
                    \resizebox{0.92\linewidth}{!}{
                    \begin{tabularx}{\columnwidth}{lccccc}
                        \toprule
                        Object & Weight (g) & Suction & Re-position & Twisting & Overall \\
                        \midrule
                        Paper & 4 &  \ding{51} &\ding{51} & \ding{51} &\ding{51 } \\
                        
                        Strawberry   & 4.6   & \ding{51}   & \ding{51}   &  \ding{51} &\ding{51}   \\
                        Pear         & 12.3  & \ding{51}  &\ding{51}   & \ding{51}  & \ding{51}  \\
                        Glue stick & 14.6 &  \ding{51} & \ding{51} & \ding{51} & \ding{51}  \\
                        Peach        & 19.9  & \ding{51}  &\ding{51}   & \ding{51}  &\ding{51}   \\
                        Lime         & 31.4  &\ding{51}   & \ding{51}  &  \ding{51} & \ding{51}  \\
                        Soft object  & 35.7  &\ding{51}   &   \ding{51}&  \ding{51} & \ding{51}  \\
                        Orange       & 74.8  & \ding{51}  & \ding{51}  & \ding{51}  &\ding{51}   \\
                        Rubik's cube & 132   & \ding{51}  &\ding{51}   &  \ding{51} &\ding{51}   \\
                        Grapes       & 139.1 & \ding{55}  &  \ding{55} &  \ding{55} & \ding{55}  \\
                        Rock         & 246.9 &\ding{55}   & \ding{55}  & \ding{55}  &   \ding{55} \\
                        Weight box & 579 & \ding{51} &  \ding{51}  & \ding{55} & \ding{55} \\
                        \bottomrule
                    \end{tabularx}}
                    \label{table-suction}
                    \vspace{-0.7cm}
                \end{table}

                The grasping sequence (top-down suction, telescopic extrusion, twisting, withdraw) was evaluated across all target objects, with the overall performance summarized in Table \ref{table-suction}. Approximately 75\% of the objects successfully adhered to the suction cup under maximum vacuum pressure (around -80 kPa). Once secured, these objects could be re-positioned and re-oriented (twisted) effectively. However, several objects failed to adhere during the initial step, leading to an overall failure in the grasping sequence by the active palm. For instance, the contact area between the suction cup and certain objects, such as the grapes and rock, was insufficient, resulting in unsealed gaps during the suction process. Consequently, all subsequent tasks were unsuccessful as they were not securely held by the suction cup from the outset.
                In contrast, the aluminum plate was successfully adhered by the suction cup; however, its size caused collisions with fingers and belts when the telescopic actuator attempted to pull it toward the gripper's center (see the Supplementary Video).

                It is important to note that the twisting motion facilitated by the suction cup is relatively simple, as it only occurs along the x-axis (in \figurename{}\ref{fig:design}), and cannot be altered until it interacts with the active surfaces. The subsequent experiment will examine the collaborative grasping and manipulation between the active palm and soft fingers with active surface.


          \subsection{ Grasping \& Manipulation}
            \begin{figure*}[t]
            \noindent
                \centering
                \includegraphics[width=\linewidth]{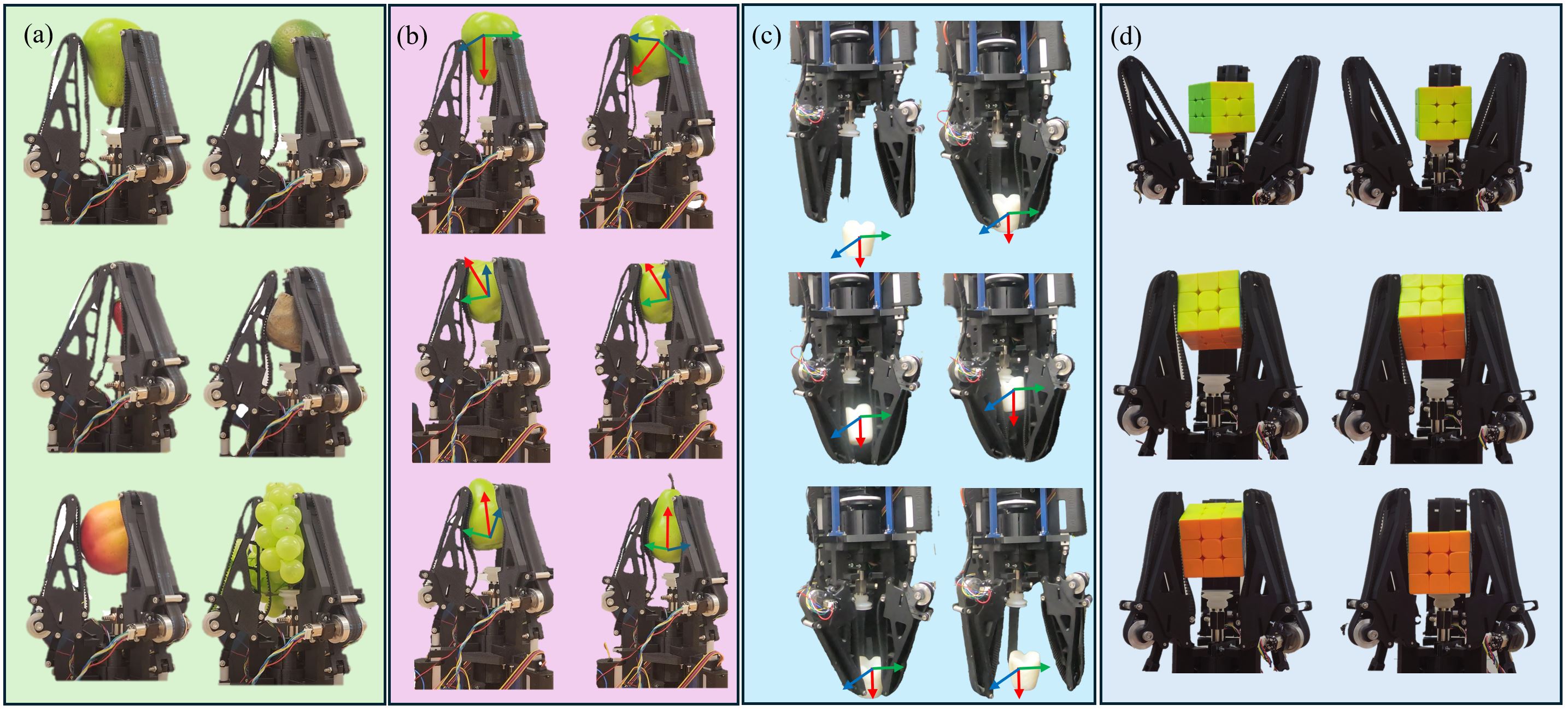}
                \caption{ (a) Functional grasping of multiple distinct objects: pear, strawberry, peach, lime, rock, grapes, (b)The gripper demonstrates the manipulation of a pear through pivot rotation: from a pinch to power grasping, (c) grasping and re-position of a deformable EGAD object through all-belt translation, (d) manipulating a Rubik's cube.}
                \vspace{-0.7cm}
                \label{combine-2}
            \end{figure*} 
             This section evaluated the DexGrip as a whole system, including its overall grasping, as well as the in-hand manipulation performance with/without the active palm.              
             
             We first conducted grasping tasks using DexGrip. In this test, the robotic end-effector was positioned stationary in an upright orientation. Objects were subsequently placed between the fingers, followed by manual closure of the gripper. The gripper was then adjusted to a grasping position, allowing the Fin Ray fingers to deform around objects. In this configuration, the stability of the grasped object was evaluated by maneuvering the robotic end-effector and visually inspecting any movements of the object.

            The results indicate that the robotic gripper can securely grasp objects ranging in size from $29.5\times32.5\times35.8$ mm to $74.2\times76.1\times79.8$ mm, with the dimensions derived from a plastic strawberry and orange respectively, as shown in \figurename{} \ref{combine-2} (a). The end-effector maintains stable grasping without significant positional shifts of the objects. The weights of the held objects vary considerably, from 4.6 g (lightest) to 246.9 g (heaviest). The maximum load capacity of the robotic end-effector is affected by the stiffness of the Fin Rays; stiffer Fin Rays provide greater grasping force.  
            However, increased stiffness may lead to greater deformation and stress concentration on delicate objects, which is undesirable for achieving non-destructive soft grasping.
             
            In the following test, the active-surface fingers were evaluated for their in-hand manipulation capabilities (transitioning from pinch to power grasp) across various objects.  
            As illustrated in \figurename{} \ref{combine-2} (b), the experimental setup involved positioning the gripper upright with suction cup facing upwards. A pear placed atop the active palm simulating a successful retraction of an object using the active palm. 
            The pinch-to-power grasp capability was assessed by having the gripper initially hold an object at the fingertip with pinch grasping, where the stem of the pear was aligned with the local x-axis (red axis).
            The pear was manipulated by controlling the rotational velocities of specific belts, enabling both its re-positioning and re-orientation. The stem was gradually rotated until its local x-axis faced upwards, completing the transition from pinch to power grasp.
            \figurename{}\ref{combine-2}(c) showed the manipulation of one deformable 3D printed object, which has a shore hardness of 40A. DexGrip was placed upside down with suction cup facing downwards. It was controlled by simultaneously driving all the belts inwards, which successfully re-positioned the object and switched from pinch to power grasping pattern. The deformable object was repositioned efficiently, requiring only minimal rotation.        

        \subsection{Collaborative Motion}         
            To further validate the gripper's ability to manipulate challenging objects beyond the usage of a purely active surface, we conducted an additional experiment coordinating the active palm and the active surface. 
            This experiment was setup with a Rubik's cube placed on top of the active palm. The bending angles of the fingers were manually adjusted to ensure sufficient contact while minimizing friction between the belts and the cube. The active palm was initially actuated to stop at an appropriate stroke, ending up with a desired in-hand position for the cube. Then the cube was rotated by the active suction cup to have one of its flat surface facing the gap between two fingers, as shown in \figurename{} \ref{combine-2} (d).
            After the engagement of the active palm, all the active surfaces were actuated in the same direction, raising the cube off the active palm, so the cube could rotate without hitting the active palm. 
            Once stationary, one of the belts would be actuated, rotating the cube as it pivoted off the two opposing fingers. This manipulation sequence could raise the cube's height. All three belts would finally actuate in the same direction, pulling the cube back to the original manipulation starting height.
            With the predefined grasping sequence, the gripper successfully re-oriented the Rubik's cube, flipping from the yellow surface to the orange one. More face flipping of the Rubik's cube can be seen in the \textbf{Supplementary Video}.


        

\section{Conclusion} \label{section: conclusion} \label{conclusion}
        This study introduces a dexterous, multi-functional soft robotics gripper capable of performing multi-modal grasping and in-hand manipulation, leveraging its collaborative multi-DoF structure. A detailed design analysis and comprehensive experimental evaluation are provided to demonstrate the effectiveness of the proposed multi-modal gripper.
        The gripper achieves conformable soft grasping, while also successfully demonstrating active adhesion, re-positioning, and re-orientation of objects with distinct surface textures and geometries using the active palm. Most notably, it can perform in-hand object manipulation by synchronizing active surface rotation, finger actuation, and active palm engagement. The integration of belt augmentation and suction cup rotation proves effective in manipulating held objects, as the belts conform to the object’s shape due to slack, and the suction cup rotation enhances operator control, ensuring efficient contact between the objects and the belts.
        Overall, these features significantly enhance the dexterity of the soft robotics gripper.     

        Further development is necessary to enhance the design. Based on the testing results, several improvements can be identified: reducing friction in the belt system, increasing the extension of the active palm, and automating the control scheme. 
        Reducing friction between the belts and the Fin Ray during rotation will enable the design to manipulate objects with greater stability. 
        

\bibliographystyle{IEEEtran}

\bibliography{root}

\end{document}